\documentclass[11pt]{article}
\usepackage{a4}
\usepackage{a4wide}
\usepackage[dvips]{graphicx}
\usepackage{xspace}
\usepackage{examples}
\usepackage{amsmath}
\usepackage{amssymb}

\renewcommand{\paragraph}[1]{\vspace{2mm}\noindent\textbf{#1}}
\newcommand{\pref}[1]{(\ref{#1})}               
\newcommand{\qit}[1]{\textsl{``#1''}}           

\newcommand{\topl}{\textsc{top}\xspace}
\newcommand{\botl}{\textsc{bot}\xspace}
\newcommand{\sqll}{\textsc{sql}\xspace}
\newcommand{\tsql}{\textsc{tsql2}\xspace}
\newcommand{\hpsg}{\textsc{hpsg}\xspace}
\newcommand{\fopl}{\textsc{fopl}\xspace}

\newcommand{\subper}{\ensuremath{\sqsubseteq}}       
\newcommand{\propsubper}{\ensuremath{\sqsubset}}     
\newcommand{\union}{\cup}                            
\newcommand{\intersect}{\cap}                        
\newcommand{\defeq}{\equiv}                          
\newcommand{\denot}[2]{\|#2\|^{#1}}                  
\newcommand{\tup}[1]{\langle#1\rangle}               

\newcommand{\partop}{\ensuremath{\mathit{Part}}\xspace}
\newcommand{\pres}{\ensuremath{\mathit{Pres}}\xspace}
\newcommand{\past}{\ensuremath{\mathit{Past}}\xspace}
\newcommand{\culm}{\ensuremath{\mathit{Culm}}\xspace}
\newcommand{\perf}{\ensuremath{\mathit{Perf}}\xspace}
\newcommand{\at}{\ensuremath{\mathit{At}}\xspace}
\newcommand{\ntense}{\ensuremath{\mathit{Ntense}}\xspace}
\newcommand{\before}{\ensuremath{\mathit{Before}}\xspace}
\newcommand{\after}{\ensuremath{\mathit{After}}\xspace}

\newcommand{\fills}{\ensuremath{\mathit{Fills}}\xspace}
\newcommand{\for}{\ensuremath{\mathit{For}}\xspace}

\newcommand{\pts}{\ensuremath{\mathit{PTS}}\xspace}
\newcommand{\periods}{\ensuremath{\mathit{PERIODS}}\xspace}

\newcommand{\cons}{\ensuremath{\mathit{CONS}}\xspace}
\newcommand{\vars}{\ensuremath{\mathit{VARS}}\xspace}
\newcommand{\terms}{\ensuremath{\mathit{TERMS}}\xspace}
\newcommand{\pfuns}{\ensuremath{\mathit{PFUNS}}\xspace}
\newcommand{\parts}{\ensuremath{\mathit{PARTS}}\xspace}
\newcommand{\cparts}{\ensuremath{\mathit{CPARTS}}\xspace}
\newcommand{\gparts}{\ensuremath{\mathit{GPARTS}}\xspace}
\newcommand{\aforms}{\ensuremath{\mathit{AFORMS}}\xspace}

\newcommand{\ynforms}{\ensuremath{\mathit{YNFORMS}}\xspace}
\newcommand{\literal}{\ensuremath{\mathit{LITERAL}}\xspace}
\newcommand{\objs}{\ensuremath{\mathit{OBJS}}\xspace}
\newcommand{\vqty}{\ensuremath{\mathit{VQTY}}\xspace}

\newcommand{\fcons}{\ensuremath{\mathit{f_{cons}}}\xspace}
\newcommand{\fpfuns}{\ensuremath{\mathit{f_{pfuns}}}\xspace}
\newcommand{\fculms}{\ensuremath{\mathit{f_{culms}}}\xspace}
\newcommand{\fgparts}{\ensuremath{\mathit{f_{gparts}}}\xspace}
\newcommand{\fcparts}{\ensuremath{\mathit{f_{cparts}}}\xspace}
\newcommand{\pow}{\ensuremath{\mathit{pow}}\xspace}
\newcommand{\mxlpers}{\ensuremath{\mathit{mxlpers}}\xspace}

\newcommand{\botsubper}{\ensuremath{\mathit{subper}}\xspace}
\newcommand{\botpart}{\ensuremath{\mathit{part}}\xspace}
\newcommand{\boteq}{\ensuremath{\mathit{eq}}\xspace}
\newcommand{\botperiod}{\ensuremath{\mathit{period}}\xspace}
\newcommand{\botintersect}{\ensuremath{\mathit{intersect}}\xspace}
\newcommand{\botprec}{\ensuremath{\mathit{prec}}\xspace}
\newcommand{\botsucc}{\ensuremath{\mathit{succ}}\xspace}
\newcommand{\botearliest}{\ensuremath{\mathit{earliest}}\xspace}
\newcommand{\botlatest}{\ensuremath{\mathit{latest}}\xspace}
\newcommand{\botbeg}{\ensuremath{\mathit{beg}}\xspace}
\newcommand{\botend}{\ensuremath{\mathit{end}}\xspace}
\newcommand{\botnow}{\ensuremath{\mathit{now}}\xspace}
\newcommand{\perex}{\ensuremath{\mathit{PEREX}}\xspace}
\newcommand{\ptex}{\ensuremath{\mathit{PTEX}}\xspace}
\newcommand{\botfpfuns}{\ensuremath{\mathit{f_{pfuns}^B}}\xspace}

\newcommand{\trans}{\mathit{trans}}

\newcommand{\customtitle}[1]{\title{\textbf{#1}}}

\customtitle{Temporal Meaning Representations in a \\ Natural Language Front-End}
\author{Ion Androutsopoulos}
\date{Software and Knowledge Engineering Laboratory \\
      Institute of Informatics and Telecommunications \\
      National Centre for Scientific Research ``Demokritos'' \\
      153 10 Ag.\ Paraskevi, Athens, Greece \\
      e-mail: \texttt{ionandr@iit.demokritos.gr}}

\begin{document}
\maketitle
\pagestyle{empty}
\thispagestyle{empty}


\begin{abstract}
Previous work in the context of natural language querying of temporal
databases has established a method to map automatically from a large subset
of English time-related questions to suitable expressions of a temporal
logic-like language, called \topl. An algorithm to translate from \topl to
the \tsql temporal database language has also been defined. This paper shows
how \topl expressions could be translated into a simpler logic-like language,
called \botl. \botl is very close to traditional first-order predicate logic
(\fopl), and hence existing methods to manipulate \fopl expressions can be
exploited to interface to time-sensitive applications other than \tsql
databases, maintaining the existing English-to-\topl mapping.
\end{abstract}


\section{Introduction} \label{introduction}

Time is an important research issue in linguistics (e.g.\ \cite{Comrie},
\cite{Comrie2}, \cite{Parsons1990}), logics (e.g.\ \cite{Gabbay1994b},
\cite{VanBenthem}), and computer systems (e.g. temporal databases
\cite{Tansel3} \cite{Tsotras1996}). In \cite{Androutsopoulos1996} and
\cite{Androutsopoulos1998a} a framework that integrates ideas from these
three areas was proposed in the context of natural language querying of
temporal databases. This framework consists of: (i) a formally defined
logic-like language, dubbed \topl, (ii) a systematic mapping from a large and
rich in temporal phenomena subset of English to \topl, based on the widely
used \hpsg grammar theory \cite{Pollard1} \cite{Pollard2}, and (iii) an
algorithm to translate from \topl to \tsql, \tsql being a recent temporal
extension of the \sqll database language that has been proposed by the
temporal databases community \cite{TSQL2book}. The framework allows written
time-related English questions to be answered automatically, by converting
them into \topl and then \tsql expressions, and executing the resulting \tsql
queries. The framework improves on previous approaches to natural language
querying of temporal databases (e.g.\ \cite{Clifford}, \cite{De2}), mainly in
terms of linguistic coverage, existence of formal definitions, and
implementation (see \cite{Androutsopoulos1996} and
\cite{Androutsopoulos1998a} for details).\footnote{A prototype implementation
of this framework is freely available from \texttt{http://www.dai.ed.ac.uk/
groups/nlp}.}

This paper shows how \topl expressions can be translated automatically into a
simpler logic-like representation language, called \botl. \botl is very close
to traditional first-order predicate logic (\fopl). Hence, existing methods
to manipulate \fopl expressions can be exploited, to interface to
time-sensitive applications other than \tsql databases (e.g.\ hybrids of
standard \sqll and Prolog databases \cite{Ceri1989} \cite{Draxler1992}
\cite{Lucas1988}, or planners \cite{Crouch2}), maintaining the existing
English-to-\topl linguistic front-end. The mapping from \topl to \botl is
also expected to make the linguistic front-end easier to interface to
forthcoming new temporal \sqll versions \cite{Snodgrass1998}, as \botl is
much simpler than \topl, and hence establishing a mapping from \botl to a new
database language is easier than from \topl.

This paper focuses on \topl, \botl, and the mapping from \topl to \botl.
Information about other aspects of the work mentioned above, including the
English-to-\topl mapping, can be found in \cite{Androutsopoulos1996} and
\cite{Androutsopoulos1998a}. The remainder of this paper is organised as
follows: Section \ref{top-language} introduces \topl, Section
\ref{bot-language} presents \botl, Section \ref{top-to-bot} describes the
\topl-to-\botl mapping, and Section \ref{conclusions} concludes. Formal
definitions of \topl and \botl can be found in Appendices
\ref{top-definitions} and \ref{bot-definitions} respectively. Appendix
\ref{top-to-bot-rules} provides a full list of the translation rules that are
used in the \topl-to-\botl mapping.


\section{The TOP language} \label{top-language}

This section introduces informally \topl, the logic-like language English
questions are initially translated into. A formal definition of the syntax
and semantics of \topl can be found in Appendix \ref{top-definitions}. \topl
was designed to support the systematic representation of English time-related
semantics, rather than inferencing (contrary to the logics of e.g.\
\cite{Allen1983}, \cite{Kowalski1986}, \cite{McCarthy1969},
\cite{McDermott1982}). Hence, although in many ways similar to traditional
temporal logics, \topl is not a full logic, as it provides no proof theory.

\topl atomic formulae are constructed by applying predicate
symbols to constants and variables. More complex formulae are
constructed using conjunctions and temporal operators (\topl is an
acronym for Temporal OPerators). For example, \pref{top:10} is
represented by the \topl formula \pref{top:20}. The ``$^v$''
suffix marks variables, and free variables (e.g.\ the $e^v$ in
\pref{top:20}) are treated as quantified by an implicit
existential quantifier with scope over the entire formula. \topl
currently provides no disjunction, negation, or explicit
quantification mechanisms, as these were not needed for the
linguistic phenomena that the work being reported here focused on.
Such mechanisms can be added easily in future \topl versions.
\begin{examples}
\item Was tank 5 empty on 1/1/98? \label{top:10}
\item $\at[1/1/98, \past[e^v, empty(tank5)]]$ \label{top:20}
\end{examples}

Roughly speaking, the verb tense in \pref{top:10} introduces a
\past operator, which requires $empty(tank5)$ to have been empty
at some past time $e^v$, and the \qit{at 1/1/98} adverbial
introduces an \at operator, which requires that past time to fall
within 1/1/98. (Unlike Priorean operators \cite{Prior}, \topl's
\past and \at operators do not shift the time where their argument
is expected to hold. They simply accumulate restrictions on what
this time can be. This is explained further below.) Assuming that
1/1/98 falls in the past, the \past operator of \pref{top:20} is
actually redundant, since any time that falls within 1/1/98 will
also belong to the past. It is important to realise, however, that
the mapping from English to \topl is carried out automatically.
This mapping introduces a \past operator when encountering the
past tense, to ensure that a sentence like \pref{top:22}, where no
adverbial is present, is represented correctly (as in
\pref{top:24}).
\begin{examples}
\item Tank 5 was empty. \label{top:22}
\item $\past[e^v, empty(tank5)]]$ \label{top:24}
\end{examples}
The combination of \at and \past operators in \pref{top:20} also
accounts for the oddity of \pref{top:10} when uttered before
1/1/98. The oddity of \pref{top:10} can be attributed to the fact
that in this case the \at and \past operators introduce
incompatible restrictions (no past time can fall within 1/1/98 if
the question is uttered before 1/1/98).

Temporal operators are used in \topl (much as in \cite{Crouch2}) to introduce
compact chunks of semantics, in a manner that makes it easier to track the
semantic contribution of each linguistic mechanism. No claim is made that
\topl is more expressive than (or even as expressive as) other temporal
representation formalisms (e.g.\ \cite{Prior}), though it should be noted
that \topl is part of a complete path from English to an application
formalism (\tsql), which is not available with most other temporal
representation formalisms.

Time in \topl is linear, discrete and bounded \cite{Gabbay1994b}
\cite{VanBenthem}. Following Reichenbach \cite{Reichenbach}, formulae are
evaluated with respect to three times: \emph{speech time} ($st$), \emph{event
time} ($et$), and \emph{localisation time} ($lt$). Intuitively, $st$ is the
time when the question is submitted, $et$ is the time when the situation of
the formula holds, and $lt$ is a temporal window that contains $et$. (\topl's
$lt$ is different from Reichenbach's reference time, and closer to the
``location time'' of \cite{Kamp1993}.) $st$ is a time-point, while $et$ and
$lt$ are generally periods. ``Period'' is used here to refer to what
logicians usually call ``intervals'', i.e.\ convex sets of time-points.

In \pref{top:10}, the answer will be affirmative if \pref{top:20} evaluates
to true. When evaluating a formula, $lt$ initially covers the entire time,
but it can be narrowed down by temporal operators. In \pref{top:20}, the \at
and \past operators narrow $lt$ to its intersection with 1/1/98 and
$[t_{first}, st)$ respectively, where $t_{first}$ is the beginning of time.
Assuming that 1/1/98 lies entirely in the past, the resulting $lt$ is 1/1/98.
The formula evaluates to true iff there is an $et$ where $empty(tank5)$ is
true, and $et \subper lt$. ($p_1$ is a subperiod of $p_2$, written $p_1
\subper p_2$ iff $p_1, p_2$ are periods and $p_1 \subseteq p_2$.)

The semantics of \topl guarantee that \topl predicates are always
\emph{homogeneous}, meaning that if a predicate is true at some $et$, it will
also be true at any other $et' \subper et$. (A similar notion of homogeneity
is used in \cite{Allen1984}.) In \pref{top:20}, if tank 5 was empty from
30/12/97 to 10/1/98 (dates are shown in the dd/mm/yy format), $empty(tank5)$
will be true at any $et$ that is a subperiod of that period. Hence, there
will be an $et$ that is a subperiod of 1/1/98 (the $lt$) where $empty(tank5)$
holds, and \pref{top:20} will evaluate to true. The reading of \pref{top:10}
that requires the tank to have been empty \emph{throughout} 1/1/98, which is
easier to grasp in the affirmative \pref{top:30}, is expressed as
\pref{top:40}. The \fills operator requires $et$ to cover the entire $lt$.
\begin{examples}
\item Tank 5 was empty on 1/1/98. \label{top:30}
\item $\at[1/1/98, \past[e^v, \fills[empty(tank5)]]]$  \label{top:40}
\end{examples}

The remainder of this section illustrates the use of some of
\topl's temporal operators, narrowing the discussion to the
representation of yes/no single-clause questions. To save space,
some of \topl's mechanisms, including those that are used to
represent wh-questions (e.g.\ \qit{Which tanks were empty on
1/1/98?}) and multiple clauses (e.g.\ \qit{Which flights were
circling while BA737 was landing?}), are not covered (see
\cite{Androutsopoulos1996} and \cite{Androutsopoulos1998a} for the
full details).

With verbs that refer to situations with inherent climaxes \cite{Moens2}
\cite{Vendler}, non-progressive tenses introduce an additional \culm
operator, which requires $et$ to be the period from the point where the
situation first started to the point where the situation last stopped, and
the situation to reach its climax at the end of $et$. For example,
\pref{top:50} and \pref{top:70} are mapped to \pref{top:60} and \pref{top:80}
respectively. \pref{top:80} requires the building to have been completed, and
the entire building to have taken place within 1997. In contrast,
\pref{top:60} simply requires part of the bulding to have been ongoing in
1997.
\begin{examples}
\item Was HouseCorp building bridge 2 in 1997? \label{top:50}
\item $\at[1997, \past[e^v, buiding(housecorp, bridge2)]]$ \label{top:60}
\item Did HouseCorp build bridge 2 in 1997? \label{top:70}
\item $\at[1997, \past[e^v, \culm[building(housecorp, bridge2)]]]$ \label{top:80}
\end{examples}

Questions referring to present situations are represented using the \pres
operator, which simply requires $st$ to fall within $et$. \pref{top:90}, for
example, is represented as \pref{top:100}.
\begin{examples}
\item Is tank 5 empty? \label{top:90}
\item $\pres[empty(tank5)]$ \label{top:100}
\end{examples}

The \perf operator is used to express the perfective aspect of questions like
\pref{top:110}. The \perf operator introduces a new event time (denoted by
$e2^v$ in \pref{top:120}) that must precede the original one ($e1^v$). In
\pref{top:120}, the inspection time ($e2^v$) must precede another past time
($e1^v$). The latter corresponds to Reichenbach's \emph{reference time}
\cite{Reichenbach}, a time that serves as a view-point.
\begin{examples}
\item Had J.~Adams inspected BA737? \label{top:110}
\item $\past[e1^v, \perf[e2^v, \culm[inspecting(jadams, ba737)]]]$
\label{top:120}
\end{examples}
In \pref{top:130}, the \qit{on 1/1/95} may refer to either the inspection
time or the reference time. The two readings are represented by
\pref{top:140} and \pref{top:150} respectively (the English to \topl mapping
generates both).
\begin{examples}
\item Had J.~Adams inspected BA737 on 1/1/95? \label{top:130}
\item $\past[e1^v, \perf[e2^v, \at[1/1/95, \culm[inspecting(jadams,ba737)]]]]$
\label{top:140}
\item $\at[1/1/95, \past[e1^v, \perf[e2^v, \culm[inspecting(jadams,ba737)]]]]$
\label{top:150}
\end{examples}

The \ntense operator (borrowed from \cite{Crouch2}) is useful in
questions like \pref{top:160}, where \qit{the president} may refer
to either the present or the 1995 president. The two readings are
captured by \pref{top:170} and \pref{top:180} respectively. In
\pref{top:180}, the $e^v$ arguments of the \past and \ntense
operators are used to ensure that both $president(p^v)$ and
$visiting(p^v, athens)$ hold at the same time.\footnote{The $e^v$
arguments of the \past and \perf operators are also used in
time-asking questions. Consult \cite{Androutsopoulos1996} and
\cite{Androutsopoulos1998a} for related discussion.}
\begin{examples}
\item Did the president visit Athens in 1995? \label{top:160}
\item $\ntense[now, president(p^v)] \land \at[1995, \past[e^v, visiting(p^v,
athens)]]$ \label{top:170}
\item $\ntense[e^v, president(p^v)] \land \at[1995, \past[e^v, visiting(p^v,
athens)]]$ \label{top:180}
\end{examples}

The reading of \pref{top:190} that asks if tank 5 was empty at
some time after 5:00 pm is represented as \pref{top:200}. The
\partop operator forces $f^v$ to range over 5:00pm-times, and the
\after operator requires the past $et$ where tank 5 is empty to
follow $f^v$.
\begin{examples}
\item Was tank 5 empty after 5:00pm? \label{top:190}
\item $\partop[\textit{5:00pm}, f^v] \land \after[f^v, \past[e^v, empty(tank5)]]$
\label{top:200}
\end{examples}
In practice, \pref{top:190} would be uttered in a context where
previous discourse has established a temporal window that contains
a single 5:00pm-time, and \qit{at 5:00pm} would refer to that
time. This anaphoric use of \qit{at 5:00pm} can be captured by
setting the initial value of $lt$ to the discourse-defined window,
rather than the entire time-axis.

Finally, durations can be specified with the \for operator. \pref{top:210} is
mapped to \pref{top:220}, which requires 45 consecutive minute-periods to
exist, and the flight to have been circling throughout the concatenation of
these periods.
\begin{examples}
\item Was BA737 circling for 45 minutes? \label{top:210}
\item $\for[minute, 45, \past[e^v, circling(ba737)]]$ \label{top:220}
\end{examples}


\section{The BOT language} \label{bot-language}

Let us now turn to \botl, the simpler formal language \topl expressions are
subsequently translated into. \botl is essentially the traditional
first-order predicate logic (\fopl), with some special terms and predicates
to refer to time-points and periods. As in \topl, \botl assumes that time is
discrete, linear, and bounded.

For simplicity, the same constant and predicate symbols are used as in the
corresponding \topl expressions. It is assumed, however, that \botl
predicates that correspond to \topl predicates have an additional argument,
whose denotations range over the maximal event-time periods where the
corresponding \topl predicates hold. For example, \pref{top:10} could be
represented in \botl as \pref{bot:20} (cf.\ \pref{top:20}).
\begin{examples}
\item $empty(tank5, p^v) \land
\botsubper(e^v, p^v) \; \land$ \\
$\botsubper(e^v, \botintersect(\botintersect([\botbeg, \botend], 1/1/98),
[\botbeg,\botnow)))$ \label{bot:20}
\end{examples}
\pref{bot:20} requires $p^v$ to denote a maximal period where the tank was
empty, and $e^v$ to be a subperiod of both $p^v$ and the intersection of
1/1/98 with the past. $e^v$ corresponds to the event time of \pref{top:20},
and $\botintersect(\botintersect([\botbeg, \botend], 1/1/98),
[\botbeg,\botnow))$ emulates \topl's localisation time, initially the
entire time-axis, which has been narrowed to cover past points within 1/1/98.
($\botbeg$, $\botend$, and $\botnow$ denote the beginning of time, end of
time, and speech time respectively, \botintersect denotes set intersection,
and square and round brackets are used to specify the boundaries of periods
in the usual manner.) As in \topl, free variables are treated as existentially
quantified.

A special \botl predicate symbol \botpart, similar to \topl's \partop
operator, allows variables to range over families of periods. In
\pref{bot:40}, for example, $m1^v$ and $m2^v$ range over minute-periods.
\botearliest and \botlatest are used to refer to the earliest and latest
time-points of a period, \botsucc steps forward one-time point, and \boteq
requires the denotations of its arguments to be identical. \pref{bot:40}
requires a 2-minute long $e^v$ period to exist, and $e^v$ to fall within the
past and be a subperiod of a maximal period $p^v$ where tank 5 was empty.
As in \pref{bot:20}, \botintersect predicates are used to emulate \topl's
localisation time (here, $\botintersect([\botbeg, \botend],
[\botbeg, \botnow))$). \pref{bot:40} represents \pref{bot:30}.
\begin{examples}
\item Was tank 5 empty for two minutes? \label{bot:30}
\item $\botpart(minute,m1^v) \; \land \; \botpart(minute,m2^v) \; \land
 \boteq(\botearliest(m1^v),\botearliest(e^v)) \; \land $ \\
$\boteq(\botsucc(\botlatest(m1^v)),\botearliest(m2^v)) \; \land \;
\boteq(\botlatest(m2^v),\botlatest(e^v)) \; \land $
\\ $empty(tank5, p^v) \;\land \; \botsubper(e^v, p^v)
\; \land \botsubper(e^v, \botintersect([\botbeg, \botend],
[\botbeg, \botnow)))$ \label{bot:40}
\end{examples}

The semantics of \botl is much simpler than \topl, though
\botl formulae tend to be much longer, and hence difficult to grasp, than the
corresponding \topl ones. The syntax and semantics of \botl are defined
formally in Appendix \ref{bot-definitions}.


\section{Translating from TOP to BOT} \label{top-to-bot}

\topl formulae are translated systematically into \botl using a set of
rewrite rules of the form:
\[ \trans(\phi_1, \varepsilon, \lambda) = \phi_2 \]
where $\phi_1$ is a \topl formula, $\phi_2$ is a \botl formula (possibly
containing recursive invocations of other translation rules), and
$\varepsilon, \lambda$ are \botl expressions representing \topl's event and
location times respectively. There are base (non-recursive) translation rules
for atomic \topl formulae, and recursive translation rules for conjunctions
and each one of \topl's operators. For example, the translation rule for
\topl's \at operator is \pref{trans:10}.
\begin{examples}
\item $\trans(\at[\tau, \phi], \varepsilon, \lambda) =
  \botperiod(\tau) \land
  \trans(\phi, \varepsilon, \botintersect(\lambda, \tau))$ \label{trans:10}
\end{examples}
The translation rules essentially express in terms of \botl constructs the
semantics of the corresponding \topl constructs. \pref{trans:10}, for
example, narrows the localisation time to the intersection of its original
value with the denotation of $\tau$, which must be a period, mirroring the
semantics of \topl's \at operator (see Appendix \ref{top-definitions}).
$\phi$ is then translated into \botl using the new value of the localisation
time.

When translating from \topl to \botl, $\lambda$ is initially set to
$[\botbeg, \botend]$, which corresponds to the initial value of \topl's
localisation time. $\varepsilon$ is set to a new variable, a variable that
has not been used in any other expression. This reflects the fact that
\topl's event time is initially allowed to be any period (see the definition
of denotation w.r.t.\ $M,st$ in Appendix \ref{top-definitions}). For example,
to compute the \botl translation of \pref{top:10}, one would invoke
\pref{trans:10} as in \pref{trans:20}, where $et^v$ is a new variable that
stands for the event time.
\begin{examples}
\item $\trans(\at[1/1/98, \past[e^v, empty(tank5)]], et^v,
[\botbeg, \botend]) = $ \\
$\botperiod(1/1/98) \land
\trans(\past[e^v, empty(tank5)], et^v,
\botintersect([\botbeg, \botend], 1/1/98)$
\label{trans:20}
\end{examples}

The translation rules for \topl's \past operator and predicates are shown in
\pref{trans:30} and \pref{trans:40} respectively. The rule for \past narrows
the localisation time to the past, and requires $\beta$ to point to the event
time. (The $\beta$ argument of \topl's \past operator is useful in
time-asking questions, which are not covered in this paper.) The rule for
predicates requires the event time to be a subperiod of both the localisation
time and of a maximal period where the predicate holds (here $\beta$ is a new
variable). These are, again, in accordance with \topl's semantics. The reader
is reminded that \botl predicates that correspond to predicates in the \topl
formula have an additional argument ($\beta$ in \pref{trans:40}), which
ranges over the maximal event-time periods where the corresponding \topl
predicate holds.
\begin{examples}
\item $\trans(\past[\beta, \phi], \varepsilon, \lambda) =
  \boteq(\beta, \varepsilon) \land
  \trans(\phi, \varepsilon, \botintersect(\lambda, [\botbeg, \botnow)))$
  \label{trans:30}
\item $\trans(\pi(\tau_1, \dots, \tau_n), \varepsilon, \lambda) =
  \botsubper(\varepsilon, \lambda) \land
  \pi(\tau_1, \dots, \tau_n, \beta) \land
  \botsubper(\varepsilon, \beta)$
  \label{trans:40}
\end{examples}

Using \pref{trans:30} and \pref{trans:40}, the right-hand side of
\pref{trans:20} becomes \pref{trans:50}. The right-hand side of
\pref{trans:50} is the final result of the translation, which is essentially
the same as the hand-crafted \pref{bot:20}. (The additional $et^v$ variable
and \botperiod predicate, do not contribute significantly in this case, but
they are needed to prove the correctness of the automatic translation.)
\begin{examples}
\item $\botperiod(1/1/98) \land \boteq(e^v, et^v) \; \land \\
    \trans(empty(tank5), et^v, \botintersect(\botintersect([\botbeg, \botend],
    1/1/98), [\botbeg, \botnow))) = $ \\
    $\botperiod(1/1/98) \land \boteq(e^v, et^v) \; \land \\
    \botsubper(et^v, \botintersect(\botintersect([\botbeg, \botend],
    1/1/98), [\botbeg, \botnow))) \; \land \\
    empty(tank5, p^v) \land \botsubper(et^v, p^v)$
    \label{trans:50}
\end{examples}

As explained in section \ref{top-language}, \topl's \culm operator requires
the event time to be the period from the point where the situation described
by \culm's argument first started to the point where it last stopped, and the
situation to reach its climax at the end of the event time. To be able to
translate \topl formulae containing \culm operators, we assume two mappings
$\eta_1$ and $\eta_2$ from predicate functors to new (unused elsewhere)
predicate functors. If $\pi(\tau_1, \dots, \tau_n)$ is a predicate in a \topl
formula, $\eta_1(\pi)(\tau_1, \dots, \tau_n)$ is a \botl predicate intended
to be true if the situation of $\pi(\tau_1, \dots, \tau_n)$ reached its
climax at the point where it last stopped. $\eta_2(\pi)(\tau_1, \dots,
\tau_n, \varepsilon)$ is another \botl predicate, where $\varepsilon$ denotes
the period from the first to the last point where the situation of
$\pi(\tau_1, \dots, \tau_n)$ was ongoing. The translation rule for \culm is
\pref{trans:60}.
\begin{examples}
\item $\trans(\culm[\pi(\tau_1, \dots, \tau_n)], \varepsilon, \lambda) =
  \botsubper(\varepsilon, \lambda) \land
  \eta_1(\pi)(\tau_1, \dots, \tau_n) \land
  \eta_2(\pi)(\tau_1, \dots, \tau_n, \varepsilon)$
  \label{trans:60}
\end{examples}

Using \pref{trans:10}, \pref{trans:30}, \pref{trans:60}, and
assuming $\eta_1(building) = cmp\_building$ and $\eta_2(building)
= max\_building$, the \topl formula of \pref{top:80} (which
represents \pref{top:70}) is translated into \botl as in
\pref{trans:70}.
\begin{examples}
\item $\trans(\at[1997, \past[e^v, \culm[building(housecorp, bridge2)]]], et^v,
[\botbeg, \botend]) = $ \\
$\botperiod(1997) \; \land$ \\
$\trans(\past[e^v, \culm[building(housecorp, bridge2)]], et^v,
\botintersect([\botbeg, \botend], 1997)) = $ \\
$\botperiod(1997) \land \boteq(e^v, et^v) \; \land$ \\
$\trans(\!\!\begin{array}[t]{l}
        \culm[building(housecorp, bridge2)], et^v, \\
        \botintersect(\botintersect([\botbeg, \botend],
        1997), [\botbeg, \botnow))) =
        \end{array}$ \\
$\botperiod(1997) \land \boteq(e^v, et^v) \; \land$ \\
$\botsubper(et^v, \botintersect(\botintersect([\botbeg, \botend],
1997), [\botbeg, \botnow))) \; \land$ \\
$cmp\_building(housecorp, bridge2) \land
max\_building(housecorp, bridge2)$
\label{trans:70}
\end{examples}
Like \pref{top:80}, \pref{trans:70} requires the period from the point
where the building first started to the point where it last stopped to fall
completely within the past and within 1997. Furthermore, the building must
have reached its completion at the end of that period.

A complete list of the \topl to \botl translation rules can be found in
Appendix \ref{top-to-bot-rules}. Although beyond the scope of this paper, a
formal proof of the correctness of the \topl to \botl translation rules could
be produced by showing that the denotations of the source \topl expressions
are identical to the denotations of the resulting \botl expressions by
induction on the syntactic complexity of the \topl expressions. The same
strategy was used in \cite{Androutsopoulos1996} to prove formally the
correctness of the \topl-to-\tsql translation rules (see
\cite{Androutsopoulos1998a} for a summary of the proof).


\section{Conclusions} \label{conclusions}

This paper has shown how an existing mapping from English to a complex
temporal meaning representation formalism (\topl) can be coupled with a
mapping from that formalism to a simpler one (\botl). The simpler formalism
is very close to traditional first-order predicate logic, making it possible
to exploit existing techniques to interface to time-sensitive applications
other than \tsql databases, while maintaining the existing linguistic
front-end.


\section*{Acknowledgements}

Part of the work reported here was carried out at the Division of Informatics
of the University of Edinburgh, with support from the Greek State
Scholarships foundation, under the supervision of Graeme Ritchie and Peter
Thanisch.


\bibliographystyle{plain}
\bibliography{biblio}

\begin{thebibliography}{10}

\bibitem{Allen1983}
J.F. Allen.
\newblock {Maintaining Knowledge about Temporal Intervals}.
\newblock {\em Communications of the ACM}, 26(11):832--843, 1983.

\bibitem{Allen1984}
J.F. Allen.
\newblock {Towards a General Theory of Action and Time}.
\newblock {\em Artificial Intelligence}, 23:123--154, 1984.

\bibitem{Androutsopoulos1996}
I.~Androutsopoulos.
\newblock {\em {A Principled Framework for Constructing Natural Language
  Interfaces to Temporal Databases}}.
\newblock PhD thesis, Department of Artificial Intelligence, University of
  Edinburgh, 1996.

\bibitem{Androutsopoulos1998a}
I.~Androutsopoulos, G.D. Ritchie, and P.~Thanisch.
\newblock {Time, Tense and Aspect in Natural Language Database Interfaces}.
\newblock {\em Natural Language Engineering}, 4(3):229--276, 1998.

\bibitem{Ceri1989}
S.~Ceri, G.~Gottlob, and G.~Wiederhold.
\newblock {Efficient Database Access from Prolog}.
\newblock {\em {IEEE Transactions on Software Engineering}}, 15(2):153--163,
  1989.

\bibitem{Clifford}
J.~Clifford.
\newblock {\em {Formal Semantics and Pragmatics for Natural Language
  Querying}}.
\newblock Cambridge Tracts in Theoretical Computer Science, Cambridge
  University Press, 1990.

\bibitem{Comrie}
B.~Comrie.
\newblock {\em {Aspect}}.
\newblock Cambridge University Press, 1976.

\bibitem{Comrie2}
B.~Comrie.
\newblock {\em {Tense}}.
\newblock Cambridge University Press, 1985.

\bibitem{Crouch2}
R.S. Crouch and S.G. Pulman.
\newblock {Time and Modality in a Natural Language Interface to a Planning
  System}.
\newblock {\em Artificial Intelligence}, 63:265--304, 1993.

\bibitem{De2}
S.~De, S.~Pan, and A.B. Whinston.
\newblock {Temporal Semantics and Natural Language Processing in a Decision
  Support System}.
\newblock {\em Information Systems}, 12(1):29--47, 1987.

\bibitem{Draxler1992}
C.~Draxler.
\newblock {\em {Accessing Relational and Higher Databases through Database Set
  Predicates in Logic Programming Languages}}.
\newblock PhD thesis, University of Zurich, 1992.

\bibitem{Gabbay1994b}
D.M. Gabbay, I.~Hodkinson, and M.~Reynolds.
\newblock {\em {Temporal Logic: Mathematical Foundations and Computational
  Aspects}}.
\newblock Oxford University Press, 1994.

\bibitem{Kamp1993}
H.~Kamp and U.~Reyle.
\newblock {\em {From Discourse to Logic}}.
\newblock Kluer Academic Publishers, 1993.

\bibitem{Kowalski1986}
R.~Kowalski and M.~Sergot.
\newblock {A Logic-based Calculus of Events}.
\newblock {\em New Generation Computing}, 4:67--95, 1986.

\bibitem{Lucas1988}
R.~Lucas.
\newblock {\em {Database Applications Using Prolog}}.
\newblock Halsted Press, 1988.

\bibitem{McCarthy1969}
J.~McCarthy and P.J. Hayes.
\newblock {Some Philosophical Problems from the Standpoint of Artificial
  Intelligence}.
\newblock In {\em Machine Intelligence 4}, pages 463--502. Edinburgh University
  Press, 1969.

\bibitem{McDermott1982}
D.~McDermott.
\newblock {A Temporal Logic for Reasoning about Processes and Plans}.
\newblock {\em Cognitive Science}, 6:101--155, 1982.

\bibitem{Moens2}
M.~Moens and M.~Steedman.
\newblock {Temporal Ontology and Temporal Reference}.
\newblock {\em Computational Linguistics}, 14(2):15--28, 1988.

\bibitem{Parsons1990}
T.~Parsons.
\newblock {\em {Events in the Semantics of English: A Study in Subatomic
  Semantics}}.
\newblock MIT Press, 1990.

\bibitem{Pollard1}
C.~Pollard and I.A. Sag.
\newblock {\em {Information-Based Syntax and Semantics -- Volume 1,
  Fundamentals}}.
\newblock Center for the Study of Language and Information, Stanford, 1987.

\bibitem{Pollard2}
C.~Pollard and I.A. Sag.
\newblock {\em {Head-Driven Phrase Structure Grammar}}.
\newblock University of Chicago Press and Center for the Study of Language and
  Information, Stanford., 1994.

\bibitem{Prior}
A.~Prior.
\newblock {\em {Past, Present and Future}}.
\newblock Oxford University Press, 1967.

\bibitem{Reichenbach}
H.~Reichenbach.
\newblock {\em {Elements of Symbolic Logic}}.
\newblock Collier-Macmillan, London, 1947.

\bibitem{TSQL2book}
R.T. Snodgrass, editor.
\newblock {\em {The TSQL2 Temporal Query Language}}. Kluwer Academic
  Publishers, 1995.

\bibitem{Snodgrass1998}
R.T. Snodgrass, M.~Boehlen, C.S. Jensen, and A.~Steiner.
\newblock {Transitioning Temporal Support in TSQL2 to SQL3}.
\newblock In O.~Etzion and S.~Sripada, editors, {\em {Temporal Databases:
  Research and Practice}}, pages 150 -- 194. Springer-Verlag, Berlin, 1998.

\bibitem{Tansel3}
A.~Tansel, J.~Clifford, S.K. Gadia, S.~Jajodia, A.~Segev, and R.T. Snodgrass.
\newblock {\em {Temporal Databases -- Theory, Design, and Implementation}}.
\newblock Benjamin/Cummings, California, 1993.

\bibitem{Tsotras1996}
V.J. Tsotras and A.~Kumar.
\newblock {Temporal Database Bibliography Update}.
\newblock {\em ACM SIGMOD Record}, 25(1), 1996.

\bibitem{VanBenthem}
J.F.A.K. van Benthem.
\newblock {\em {The Logic of Time}}.
\newblock D.~Reidel Publishing Company, Dordrecht, Holland, 1983.

\bibitem{Vendler}
Z.~Vendler.
\newblock {Verbs and Times}.
\newblock In {\em Linguistics in Philosophy}, chapter~4, pages 97--121. Cornell
  University Press, Ithaca, NY, 1967.

\end{thebibliography}
\appendix
\newpage
\section*{Appendix}


\section{Definition of the TOP language} \label{top-definitions}

This section defines the syntax and semantics of the subset of \topl that
was introduced in this paper. See \cite{Androutsopoulos1996} and
\cite{Androutsopoulos1998a} for a full definition of \topl.

\subsection{Syntax of TOP} \label{top-syntax}

The syntax of \topl is defined below using \textsc{bnf}. Angle brackets are
used to group \textsc{bnf} elements. ``$^{*}$'' denotes zero or more
repetitions. ``$^{+}$'' denotes one or more repetitions. Terminal symbols are
in lower case, possibly with an initial capital. Non-terminals are in upper
case. The distinguished symbol is \ynforms.
\begin{align*}
\ynforms \rightarrow \; & \aforms \; \mid \; \ynforms \land \ynforms \\
         \mid \; & \pres[\ynforms] \; \mid \; \past[\vars, \ynforms] \;
            \mid \; \perf[\vars,\ynforms] \\
         \mid \; & \culm[\literal] \; \mid \; \at[\terms,\ynforms] \\
         \mid \; & \before[\terms,\ynforms] \; \mid \; \after[\terms,\ynforms] \\
         \mid \; & \ntense[\vars, \ynforms] \; \mid \; \ntense[\mathit{now}, \ynforms] \\
         \mid \; & \for[\cparts, \vqty, \ynforms] \; \mid \; \fills[\ynforms]
\\ \aforms  \rightarrow \; & \literal \; \mid \; \partop[\parts,\vars]
\\ \literal \rightarrow \; & \pfuns(\{\terms ,\}^* \terms)
\\ \terms   \rightarrow \; & \cons \; \mid \; \vars
\\ \parts   \rightarrow \; & \cparts \; \mid \; \gparts
\\ \vqty    \rightarrow \; & 1 \; \mid \; 2 \; \mid \; 3 \; \mid \; \ldots
\end{align*}
\noindent \pfuns, \cparts, \gparts, \cons, and \vars are disjoint open classes of
terminal symbols.

\subsection{Semantics of TOP}

\subsubsection*{Temporal ontology}\label{temporal_ontology}

A \emph{point structure} $\tup{\pts, \prec}$ is assumed, where \pts is the
set of time-points, and $\prec$ is a binary, transitive, irreflexive relation
over $\pts \times \pts$. Time is assumed to be discrete, bounded, and linear
\cite{Gabbay1994b} \cite{VanBenthem}. $t_{first}$ and $t_{last}$ are the
earliest and latest time-points respectively. $prev(t)$ and $next(t)$ are
used to refer to the immediately previous and following time-points of a $t
\in \pts$. For $S \subseteq \pts$, $minpt(S)$ and $maxpt(S)$ denote the
earliest and latest time-points in $S$.

A \emph{period} $p$ over $\tup{\pts, \prec}$ is a non-empty subset of \pts.
Periods are convex, i.e.\ if $t_1, t_2 \in p$, $t_3 \in \pts$, and $t_1 \prec
t_3 \prec t_2$, then $t_3 \in p$. \periods is the set of all periods over
$\tup{\pts, \prec}$. $p_1$ is a \emph{subperiod} of $p_2$ (written $p_1
\subper p_2$), iff $p_1, p_2 \in \periods$ and $p_1 \subseteq p_2$. $p_1$ is
a proper \emph{subperiod} of $p_2$ (written $p_1 \propsubper p_2$), iff $p_1,
p_2 \in \periods$ and $p_1 \subset p_2$. The usual notational conventions
apply when specifying the boundaries of periods; e.g. $(t_1, t_2]$ is an
abbreviation for $\{t \in \pts \mid t_1 \prec t \preceq t_2 \}$.

If $S$ is a set of periods, then $\mxlpers(S)$ is the set of \emph{maximal
periods} of $S$.  $\mxlpers(S) \defeq \{p \in S \mid \text{for no } p' \in S
\text{ is it true that } p \propsubper p'\}$.

\subsubsection*{TOP model}
\label{top_model}

A \topl model $M$ is an ordered 7-tuple:
\[ M = \tup{\tup{\pts, \prec}, \objs,
            \fcons, \fpfuns, \fculms, \fgparts, \fcparts}
\]
where $\tup{\pts, \prec}$ is the point structure, $\periods \subseteq \objs$,
and \fcons, \fpfuns, \fculms, \fgparts, and \fcparts are as specified below:

\objs is a set containing all the objects in the modelled world that can be
denoted by \topl terms, and \fcons is a function $\cons \mapsto \objs$.
Intuitively, \fcons maps each constant to the object it denotes.

\fpfuns maps each $\pi \in \pfuns$ to a function $(\objs)^n
\mapsto \pow(\periods)$. It is assumed that each predicate symbol
$\pi \in \pfuns$ is used with a particular arity (number of
arguments) $n$. $\pow(S)$ denotes the powerset (set of all
subsets) of S. For every $\pi \in \pfuns$ and every
$\tup{o_1,o_2,\dots,o_n} \in (\objs)^n$, it must be true that:
\[
\text{if } p_1, p_2 \in \fpfuns(\pi)(o_1, o_2, \dots, o_n) \text{ and }
p_1 \union p_2 \in \periods, \text{ then } p_1 = p_2
\]
Intuitively, \fpfuns shows the maximal periods where the
situation represented by $\pi(\tau_1, \dots, \tau_n)$ holds.

\fculms is a function that maps each $\pi \in \pfuns$ to a function
$(\objs)^n \mapsto \{T,F\}$. Intuitively, \fculms shows whether or not a
situation reaches a climax at the latest time-point where it is ongoing.

\fgparts is a function that maps each element of \gparts to a
\emph{gappy partitioning}. A gappy partitioning is a subset $S$ of
\periods, such that for every $p_1, p_2 \in S$, $p_1 \intersect p_2 =
\emptyset$, and $\bigcup_{p \in S}p \not= \pts$.
\fcparts is a function that maps each element of \cparts to a
\emph{complete partitioning}. A complete partitioning is a subset
$S$ of \periods, such that for every $p_1, p_2 \in S$, $p_1 \intersect
p_2 = \emptyset$, and $\bigcup_{p \in S}p = \pts$.

\subsubsection*{Variable assignment} \label{var-assign}

A variable assignment w.r.t. a \topl model $M$ is a function
$g: \vars \mapsto \objs$. $G_M$, or simply $G$,
is the set of all possible variable assignments w.r.t.\ $M$.

\subsubsection*{TOP denotation w.r.t.\ M, st, et, lt, g} \label{denotation}

Non-terminal symbols of the \topl \textsc{bnf} are used here as names of sets
that contain expressions which can be analysed syntactically as the
corresponding non-terminals.

An \emph{index of evaluation} is an ordered 3-tuple
$\tup{st,et,lt}$, such that $st \in \pts$, $et \in \periods$, and $lt
\in \periods \union \{\emptyset\}$.

The \emph{denotation} of a \topl expression $\xi$ w.r.t.\ a model $M$,
an index of evaluation $\tup{st,et,lt}$, and a variable assignment $g$,
is written $\denot{M,st,et,lt,g}{\xi}$ or simply
$\denot{st,et,lt,g}{\xi}$. When the denotation of $\xi$ does not
depend on $st$, $et$, and $lt$, we may write $\denot{M,g}{\xi}$
or simply $\denot{g}{\xi}$.
\begin{itemize}

\item If $\kappa \in \cons$, then $\denot{g}{\kappa} = \fcons(\kappa)$.

\item If $\beta \in \vars$, then $\denot{g}{\beta} = g(\beta)$.

\item If $\phi \in \ynforms$, then $\denot{st,et,lt,g}{\phi} \in
\{T,F\}$.

\item If $\pi(\tau_1, \tau_2, \dots, \tau_n) \in \literal$, then
$\denot{st,et,lt,g}{\pi(\tau_1, \tau_2, \dots, \tau_n)} = T$
iff $et \subper lt$ and for some $p_{mxl} \in
\fpfuns(\pi)(\denot{g}{\tau_1}, \denot{g}{\tau_2}, \dots, \denot{g}{\tau_n})$,
$et \subper p_{mxl}$.

\item If $\phi_1, \phi_2 \in \ynforms$, then
$\denot{st,et,lt,g}{\phi_1 \land \phi_2} = T$ iff
$\denot{st,et,lt,g}{\phi_1} = \denot{st,et,lt,g}{\phi_2} = T$.

\item$\denot{g}{\partop[\sigma, \beta]} = T$ iff $g(\beta)
\in f(\sigma)$ (where $f = \fcparts$ if $\sigma \in \cparts$, and $f
= \fgparts$ if $\sigma \in \gparts$).

\item $\denot{st,et,lt,g}{\pres[\phi]} = T$, iff $st \in et$ and
  $\denot{st,et,lt,g}{\phi} = T$.

\item $\denot{st,et,lt,g}{\past[\beta, \phi]} = T$, iff
  $g(\beta) = et$ and
  $\denot{st,et, lt \intersect [t_{first}, st), g}{\phi} = T$.

\item
$\denot{st,et,lt,g}{\culm[\pi(\tau_1, \dots, \tau_n)]} = T$, iff $et \subper
lt$, $\fculms(\pi)(\denot{g}{\tau_1}, \dots, \denot{g}{\tau_n}) = T$, $S
\not= \emptyset$, and $et = [minpt(S), maxpt(S)]$, where:
\[
S = \bigcup_{p \in \fpfuns(\pi)(\denot{g}{\tau_1}, \dots,
             \denot{g}{\tau_n})}p
\]

\item
   $\denot{st,et,lt,g}{\at[\tau, \phi]} = T$, iff
   $\denot{g}{\tau} \in \periods$ and
   $\denot{st,et,lt \intersect \denot{g}{\tau},g}{\phi} = T$.

\item
   $\denot{st,et,lt,g}{\before[\tau, \phi]} = T$, iff
   $\denot{g}{\tau} \in \periods$ and
   $\denot{st,et,
           lt \intersect [t_{first}, minpt(\denot{g}{\tau})),
           g}{\phi} = T$.

\item
   $\denot{st,et,lt,g}{\after[\tau, \phi]} = T$, iff
   $\denot{g}{\tau} \in \periods$ and
   $\denot{st,et,
           lt \intersect (maxpt(\denot{g}{\tau}), t_{last}],
           g}{\phi} = T$.

\item
   $\denot{st,et,lt,g}{\fills[\phi]} = T$, iff $et = lt$ and
   $\denot{st,et,lt,g}{\phi} = T$.

\item
  $\denot{st,et,lt,g}{\ntense[\beta, \phi]} = T$, iff for some
  $et' \in \periods$, $g(\beta)= et'$ and\\
  $\denot{st,et',\pts,g}{\phi} = T$.

\item
  $\denot{st,et,lt,g}{\ntense[now^*, \phi]} = T$, iff
  $\denot{st,\{st\},\pts,g}{\phi} = T$.

\item
$\denot{st,et,lt,g}{\for[\sigma_c, \nu_{qty}, \phi]} = T$, iff
$\denot{st,et,lt,g}{\phi} = T$, and for some
$p_1,p_2,\dots,p_{\nu_{qty}} \in \fcparts(\sigma_c)$, it is true
that $minpt(p_1) = minpt(et)$, $next(maxpt(p_1)) = minpt(p_2)$,
$next(maxpt(p_2)) = minpt(p_3)$, \dots, $next(maxpt(p_{\nu_{qty} -
1})) = minpt(p_{\nu_{qty}})$, and \\$maxpt(p_{\nu_{qty}}) =
maxpt(et)$.

\item
$\denot{st,et,lt,g}{\perf[\beta, \phi]} = T$, iff $et \subper
lt$, and for some $et' \in \periods$, it is true that $g(\beta) =
et'$, $maxpt(et') \prec minpt(et)$, and $\denot{st,et',\pts,g}{\phi}
= T$.
\end{itemize}

\subsection*{TOP denotation w.r.t.\ M, st}

The denotation of $\phi$ w.r.t.\ $M, st$, written $\denot{M,st}{\phi}$
or simply $\denot{st}{\phi}$, is defined only for $\phi \in \ynforms$:
\begin{itemize}
\item If $\phi \in \ynforms$, then $\denot{st}{\phi} =$
    \begin{itemize}
    \item $T$, if for some $g \in G$ and $et \in \periods$,
    $\denot{st,et,\pts,g}{\phi} = T$,
    \item $F$, otherwise
    \end{itemize}
\end{itemize}


\section{Definition of the BOT language} \label{bot-definitions}

\subsection{Syntax of BOT}

The syntax of \botl is defined using \textsc{bnf}, with the same conventions
as in the definition of \topl. The distinguished symbol is $\ynforms^B$.
\begin{align*}
\ynforms^B \rightarrow \; & \aforms^B \; \mid \; \ynforms^B \land \ynforms^B \\
\aforms^B  \rightarrow \; & \literal^B \; \mid \; \botsubper(\perex, \perex)
                          \; \mid \; \boteq(\terms^B, \terms^B) \\
                          & \mid \; \botperiod(\terms^B) \; \mid
                          \botpart(\parts, \terms^B) \\
\literal^B \rightarrow \; & \pfuns(\{\terms^B ,\}^* \terms^B) \\
\terms^B   \rightarrow \; & \cons \; \mid \; \vars \; \mid \; \perex \;
                          \mid \; \ptex \\
\parts     \rightarrow \; & \cparts \; \mid \; \gparts \\
\ptex    \rightarrow \; & \botbeg \; \mid \; \botnow \; \mid \; \botend \;
                          \mid \; \botearliest(\perex) \; \mid \;
                          \botlatest(\perex)\\
                          & \mid \botsucc(\ptex) \; \mid \;
                          \botprec(\ptex, \ptex)\\
\perex   \rightarrow \; & [\ptex, \ptex] \; \mid \; [\ptex, \ptex) \;
                          \mid (\ptex, \ptex] \\
                          & \mid \; (\ptex, \ptex) \; \mid
                          \botintersect(\perex, \perex)
\end{align*}
\noindent \pfuns, \cparts, \gparts, \cons, and \vars are disjoint
open classes of terminal symbols that do not contain any of the other \botl
terminal symbols. The same symbols for \pfuns, \parts, \cparts, \gparts,
\cons, \vars are used as in the definition of \topl, because these classes are
the same in both languages.

\subsection{Semantics of BOT}

\botl assumes the same temporal ontology as \topl.

\subsubsection*{BOT model}

A \botl model $M$ is an ordered 6-tuple:
\[ M^B = \tup{\tup{\pts, \prec}, \objs,
            \fcons, \botfpfuns, \fgparts, \fcparts}
\]
where $\tup{\pts, \prec}$ is the point structure, and \objs, \fcons, \fgparts,
and \fcparts is are as in \topl. $\botfpfuns$ is as a function that maps
every $\pi \in \pfuns$ to a function $\objs^n \mapsto \{T,F\}$, where $n$ is
the arity of $\pi$.

\subsubsection*{Variable assignment}

A variable assignment for \botl is a function $g: \vars \mapsto \objs$, as in
\topl. $G$ has the same meaning as in \topl.

\subsubsection*{BOT denotation w.r.t.\ M, st, g}

The denotation of a \botl expression $\xi$ w.r.t.\ a \botl model $M^B$, a speech
time $st \in \pts$, and a $g \in G$, written $\denot{M^B,st,g}{\xi}$ or simply
$\denot{st,g}{\xi}$, is defined as follows:

\begin{itemize}

\item If $\kappa \in \cons$, then $\denot{st,g}{\kappa} = \fcons(\kappa)$.

\item $\denot{st,g}{\botbeg} = t_{first}$,
$\denot{st,g}{\botnow} = st$, $\denot{st,g}{\botend} = t_{last}$.

\item $\denot{st,g}{\botearliest(\tau)} = minpt(\denot{st,g}{\tau})$.
$\denot{st,g}{\botlatest(\tau)} = maxpt(\denot{st,g}{\tau})$.

\item $\denot{st,g}{[\xi_1, \xi_2)} =
\{t \in \pts \; \mid \; \denot{st,g}{\xi_1} \preceq t \prec
\denot{st,g}{\xi_2} \}$. The denotations of $[\xi_1, \xi_2]$, $(\xi_1,
\xi_2]$, and $(\xi_1, \xi_2)$ are defined similarly.

\item $\denot{st,g}{\botintersect(\xi_1, \xi_2)} =
\denot{st,g}{\xi_1} \intersect \denot{st,g}{\xi_2}$.

\item $\denot{st,g}{\botsucc(\xi)} = next(\denot{st,g}{\xi})$.

\item $\denot{st,g}{\botprec(\xi_1, \xi_2)}$ is $T$ if
$\denot{st,g}{\xi_1} \prec \denot{st,g}{\xi_2}$ and $F$ otherwise.

\item If $\beta \in \vars$, then $\denot{st,g}{\beta} = g(\beta)$.

\item If $\phi \in \ynforms^B$, then $\denot{st,g}{\phi} \in \{T,F\}$.

\item If $\pi(\tau_1, \tau_2, \dots, \tau_n) \in \literal^B$, then
$\denot{st,g}{\pi(\tau_1, \tau_2, \dots, \tau_n)} =$ \\
$\botfpfuns(\pi)(\denot{st,g}{\tau_1}, \denot{st,g}{\tau_2}, \dots,
\denot{st,g}{\tau_n})$.

\item If $\phi_1, \phi_2 \in \ynforms^B$, then
$\denot{st,g}{\phi_1 \land \phi_2} = T$ iff
$\denot{st,g}{\phi_1} = T$ and $\denot{st,g}{\phi_2} = T$.

\item $\denot{st,g}{part(\sigma, \beta)} = T$ iff
$g(\beta) \in f(\sigma)$ (where $f = \fcparts$ if $\sigma \in \cparts$, and
$f = \fgparts$ if $\sigma \in \gparts$).

\item $\denot{st,g}{eq(\tau_1, \tau2)} = T$ iff
$\denot{st,g}{\tau_1} = \denot{st,g}{\tau_2}$.

\item $\denot{st,g}{subper(\xi_1,\xi_2)} = T$ iff
$\denot{st,g}{\xi_1} \subper \denot{st,g}{\xi_2}$.

\item $\denot{st,g}{\botperiod(\tau)} = T$ iff
$\denot{st,g}{\tau} \in \periods$.

\end{itemize}

\subsubsection*{BOT denotation w.r.t.\ M, st}

The denotation of $\phi$ w.r.t.\ $M^B, st$, written $\denot{M^B,st}{\phi}$
or simply $\denot{st}{\phi}$, is defined only for $\phi \in \ynforms^B$:
\begin{itemize}
\item If $\phi \in \ynforms^B$, then $\denot{st}{\phi} =$
    \begin{itemize}
    \item $T$, if for some $g \in G$, $\denot{st,g}{\phi} = T$,
    \item $F$, otherwise
    \end{itemize}
\end{itemize}


\section{TOP to BOT translation rules} \label{top-to-bot-rules}

\begin{itemize}
  \item If $\pi \in \pfuns$ and $\tau_1, \dots, \tau_n \in \terms$, then: \\
  $\trans(\pi(\tau_1, \dots, \tau_n), \varepsilon, \lambda) =
  \botsubper(\varepsilon, \lambda) \land
  \pi(\tau_1, \dots, \tau_n, \beta) \land
  \botsubper(\varepsilon, \beta)$, \\
  where $\beta$ is a new variable.

  \item $\trans(\phi_1 \land \phi_2, \varepsilon, \lambda) =
  \trans(\phi_1, \varepsilon, \lambda) \land
  \trans(\phi_2, \varepsilon, \lambda)$.

  \item $\trans(\partop[\sigma, \beta], \varepsilon, \lambda) =
  \botpart(\sigma, \beta)$.

  \item $\trans(\pres[\phi], \varepsilon, \lambda) =
  \botsubper([now, now], \varepsilon) \land
  \trans(\phi, \varepsilon, \lambda)$.

  \item $\trans(\past[\beta, \phi], \varepsilon, \lambda) =
  \boteq(\beta, \varepsilon) \land
  \trans(\phi, \varepsilon, \botintersect(\lambda, [\botbeg, \botnow)))$.

  \item $\trans(\culm[\pi(\tau_1, \dots, \tau_n)], \varepsilon, \lambda) =
  \botsubper(\varepsilon, \lambda) \land
  \eta_1(\pi)(\tau_1, \dots, \tau_n) \land
  \eta_2(\pi)(\tau_1, \dots, \tau_n, \varepsilon)$, \\
  where $\eta_1, \eta_2$ are as in section \ref{top-to-bot}.

  \item $\trans(\at[\tau, \phi], \varepsilon, \lambda) =
  \botperiod(\tau) \land
  \trans(\phi, \varepsilon, \botintersect(\lambda, \tau))$.

  \item $\trans(\before[\tau, \phi], \varepsilon, \lambda) =
  \botperiod(\tau) \land
  \trans(\phi, \varepsilon, \botintersect(\lambda, [\botbeg, minpt(\tau))))$.

  \item $\trans(\after[\tau, \phi], \varepsilon, \lambda) =
  \botperiod(\tau) \land
  \trans(\phi, \varepsilon, \botintersect(\lambda, (maxpt(\tau), \botend]))$.

  \item $\trans(\fills[\phi], \varepsilon, \lambda) =
  \boteq(\varepsilon, \lambda) \land \trans(\phi, \varepsilon, \lambda)$.

  \item $\trans(\ntense[\beta, \phi], \varepsilon, \lambda) =
  \botperiod(\beta) \land \trans(\phi, \beta, [\botbeg, \botend])$.

  \item $\trans(\ntense[\botnow, \phi], \varepsilon, \lambda) =
  \trans(\phi, [\botnow, \botnow], [\botbeg, \botend])$.

  \item $\trans(\for[\sigma_c, \nu_{qty}, \phi], \varepsilon, \lambda) = $ \\
  $\botpart(\sigma_c, \beta_1) \land
  \botpart(\sigma_c, \beta_2) \land \dots \land
  \botpart(\sigma_c, \beta_{\nu_{qty}}) \land $ \\
  $\boteq(\botearliest(\beta_1), \botearliest(\varepsilon)) \; \land \;
  \boteq(\botsucc(\botlatest(\beta_1)), \botearliest(\beta_2)) \; \land $ \\
  $\boteq(\botsucc(\botlatest(\beta_2)), \botearliest(\beta_3)) \; \land \;
  \dots \land \;
  \boteq(\botsucc(\botlatest(\beta_{\nu_{qty}-1})),
  \botearliest(\beta_{\nu_{qty}}) \; \land$ \\
  $\boteq(\botlatest(\beta_{\nu_{qty}}), \botlatest(\varepsilon))
  \; \land \; \trans(\phi, \varepsilon, \lambda)$,
  where $\beta_1, \beta_2, \dots, \beta_{\nu_{qty}}$ are new variables.

  \item $\trans(\perf[\beta, \phi], \varepsilon, \lambda) =$ \\
  $\botsubper(\varepsilon, \lambda) \land \botperiod(\beta) \land
  \botprec(\botlatest(\beta), \botearliest(\varepsilon)) \land
  \trans(\phi, \beta, [\botbeg, \botend])$.

\end{itemize}


\end{document}